# Defect Detection Techniques for Airbag Production Sewing Stages

**Raluca Brad[1], Lavinia Barac[2], Remus Brad[2]**

[1] Industrial Machinery and Equipment Department, Lucian Blaga University of Sibiu, B-dul Victoriei 10, 550024 Sibiu Romania, raluca.brad@ulbsibiu.ro

[2] Computer Science and Electrical Engineering Department, Lucian Blaga University of Sibiu, B-dul Victoriei 10, 550024 Sibiu Romania

**Abstract:** Airbags are subject to strict quality control in order to ensure passengers safety. The quality of fabric and sewing thread influence the final product and therefore, sewing defects must be early and accurately detected, in order to remove the item from production. Airbag seams assembly can take various forms, using linear and circle primitives, with threads of different colors and length densities, creating lockstitch or double threads chainstitch. The paper presents a framework for the automatic detection of defects occurring during the airbag sewing stage. Types of defects as skipped stitch, missed stitch or superimposed seam for lockstitch and two threads chainstitch are detected and marked. Using image processing methods, the proposed framework follows the seams path and determines if a color pattern of the considered stitches is valid.

**Keywords:** seam, defect, control, airbag, image processing

## 1. Introduction

The term of product quality refers primarily to the product ability to fulfill the purpose for which it was designed, while performing a quality control assumes a regular check of quality products variables. Therefore, the quality features which will be subsequently tested need to be defined. The textiles products are assessed according to several criteria, depending on their destination. For instance, technical textiles have to meet close specifications of weight, flexibility and strength, while in the case of apparel, quality characteristics as design, color and comfort are important.

The airbag is a safety device designed for modern vehicles, required to inflate quickly in the case of a collision, preventing the passengers from crashing onto the cars interior [1]. The concept of airbag was introduced in the late 60's, but it was increasingly employed after 1990, due to market requirements and imposed legislation. Generally, the air cushion is woven from nylon filament yarns, suitable due to the high strength-to-weight proportion, favorable elongation, adequate thermal properties, and relatively low cost of production. In average, to produce a driver's airbag, about 1.42 meters of material is needed.

Other mandatory features for airbag are high tensile strength, seam stability, air permeability and ability to be folded and stored in dark for years, without suffering any damage [2]. To fulfill the designed functionality, it is necessary that all these properties are valid. Therefore, the quality control of airbag production is extremely important, because a minor defect can influence the cushions performance. Quality control is very strict for airbags, especially in the areas of propulsion, static and dynamic inflation and of fabrics and seams quality.

In the case of fabric quality, critical parameters include tensile strength and elongation of yarn used for weaving, the tensile strength of the material, fabric density and weight, air permeability (both for static and dynamic air) and slip resistance of the seam [3]. Thus, a rigorous testing and quality control starting from the yarns used in weaving to the final sewing operation is absolutely necessary.

In the past decades, seams quality control was visually performed by human experts pursuing two goals: a continuous control to prevent the sale of poor quality products and setting the optimal parameters for sewing machines. One of the first automated techniques for seam control was reported in [4], employing a laser to scan the stitches. Also, piezoelectric or early vision methods were presented in the referenced paper. Among the electro-mechanical detections systems for irregular formed stitches, the counting of thread consumption and tension is presented in [5], [6].

In the area of image processing, an automatic detection of stitch defects was presented in [7], where two classes have been identified: seam and cloth defects. The texture features have been used to classify the defects. Another contribution is presented in [8], where the seam line is detected using the Radon transform, offering a robust recognition: invariant to illumination, motion and noise. While trying to automate the control process, Bahlmann et al. [9] proposed a method using neural networks for the detection and seams classification. The solution consists of a features extraction using a Fourier spectrum and a neural network training stage. The Fourier spectrum will reflect the fabric puckering due to a tight stitch. The Hough transform for line detection is used in the paper of Yuen et al. [10] for stitch inspection. The difficulty of the method lies in the binarization of acquired images before the Hough detection. The authors of [11] offer a solution for seams assessment employing a feature extraction and a self-organizing map. In this case, the seam pucker defect is emphasis, as another major nonconformity caused by uneven thread tension or material feed. The applications of neural networks in the field of textile are subject to a comprehensive review paper in [12].

To prevent the spread of sewing defects in the subsequent stages, a visual inspection point is inserted before achieving the final closure seam. In this checkpoint, the fixing stitches of the 3 reinforcement and the upper and lower panels' assembly are inspected. Our application can successfully perform sewing defect detection, considering the seams paths and detecting defects such as threads breakage, stitch skipping, double stitching, dimensional nonconforming, uneven stitch or missing stitch.

The paper is organized as follows: section 2 presents the materials implied in the airbag sewing stage, together with seams types and possible defects. In section 3, the image processing stages for nonconformities detection are described, while the results of linear and circular seam inspection are shown in section 4, followed by conclusions.

## 2. Seam nonconformities in airbag assembly stage

The material that fulfills the constructive function of the product consists of a plain fabric of 6.6 polyamide coated with thin film of silicon. During the weaving process, the following specifications have to be satisfied: a density of 46x46 ends and picks/inch, 470 dtex yarn count, a total of 68 filaments S twisted per yarn. As soon as a roll was removed from the loom, the fabric is coated in order to acquire certain properties: an improved abrasion resistance, a better processability in cut and sewn technology and controlled air porosity [13]. The material follows the technological flow and get to the cutting stage, after then parts of the product are assembled using the sewing threads.

Assembling the components of a passenger airbag is accomplished with seams that have different paths depending on the part shape and can be straight or circular (Table 1). In order to be accurate, sewing patterns are used during the process; however 0 defects are not achievable.

Table 1. Stitch types in airbag sewing process [14], [15], [16]

| Stitch type | Seam class | Seam line | Sewing operation |
|---|---|---|---|
| 301 lockstitch | SSa -1 / 1.01 | straight | 2 reinforcements sewing on the lower panel |
| | LSbj – 1/ 5.30.01 | circular | Vent reinforcements sewing on the lower panel |
| | SSv – 1/ 5.01 | circular | Generator reinforcements and cover sewing on the lower panel subassembly |
| 401 two threads chainstitch | SSa -1 / 1.01 | straight | The bonding sewing of the ends of the upper and lower panels |
| | SSa -1 / 1.01 | straight | Perimeter sewing |
| | SSa -1 / 1.01 | straight | Reinforcement sewing on the bottom panel |
| | SSa -2 / 1.01 | straight | The closure sewing |

Two types of stitches are being used: the 301 type lockstitch [14], a resistant stitch which stiffens more layers and the 401 type double-locked chain stitch [14], imposing a seam with high elasticity and strength, which is usually placed where shock is applied. However, compared with the lockstitch, the chain stitch is easier frayed. The lockstitch is formed by a red needle thread passing through the fabric and interlocking with the green bobbin thread, used for reinforcement grip seam on the bottom, vent and generator reinforcements seam. The chain stitch is obtained with the needle red thread and the orange looper thread, being used for perimeter seam and bonding of panels seam.

The failure of a component, such as the seams, can result in failure of the whole product, regardless the quality of other components or fabrics used in the product. Factors like wear or break of sewing threads before the fabric degradation, punching of fabric during the sewing process and a low slip resistance of seam can lead to a faulty seam. The potential causes can be controlled by the sewing variables: type of sewing machine, size of sewing needle, sewing speed, sewing thread type, and stitch length [17]. The problems caused by faulty seams could be avoided by a novel technology applying a laser welding process as an alternative joining technique of PA or PES fabrics [18].

Whatever the stitch type, a sewing defect is an unaesthetic appearance of a joint as a result of incorrect stitching and will produce one of the following nonconformities:

§ Seam puckering - a local defect appearing as material wrinkling near the stitch
§ Open seam - undesirable gap in a seam caused by thread breakage or fabric yarn breakage
§ Seam slippage – fabric yarns loosened from the seam by pulling or pressing
§ Stitch skipping – stitches are incompletely formed or skipped during the assembly

Seam appearance and performance are influenced by thread features like type, structure, twist, ply, color matching, finishes and size [19], [20], [21]. Due to a specific requirement of durability and abrasion resistance, nylon twisted multi-filament thread is used in assembling the passenger airbag parts (top panel, lower panel, reinforcements, and cover). For a reliable detection and strength seam monitoring, sewing threads of different colors and linear density are used:

green thread 1401 dtex, orange thread 932 dtex and red thread 1401 dtex. In the most important points, which directly affect the functionality of the airbag, less fine threads with high tensile strength are employed.

## 3. A method for airbag seam inspection

In our practical approach, we have considered two types of seam paths: linear and circular or partial circular (arcs). Three sewing nonconformities have been considered for detection: missing stitches, skipped stitches and superimposed seam. The corresponding image processing steps for each class of defects are presented in figure 1.

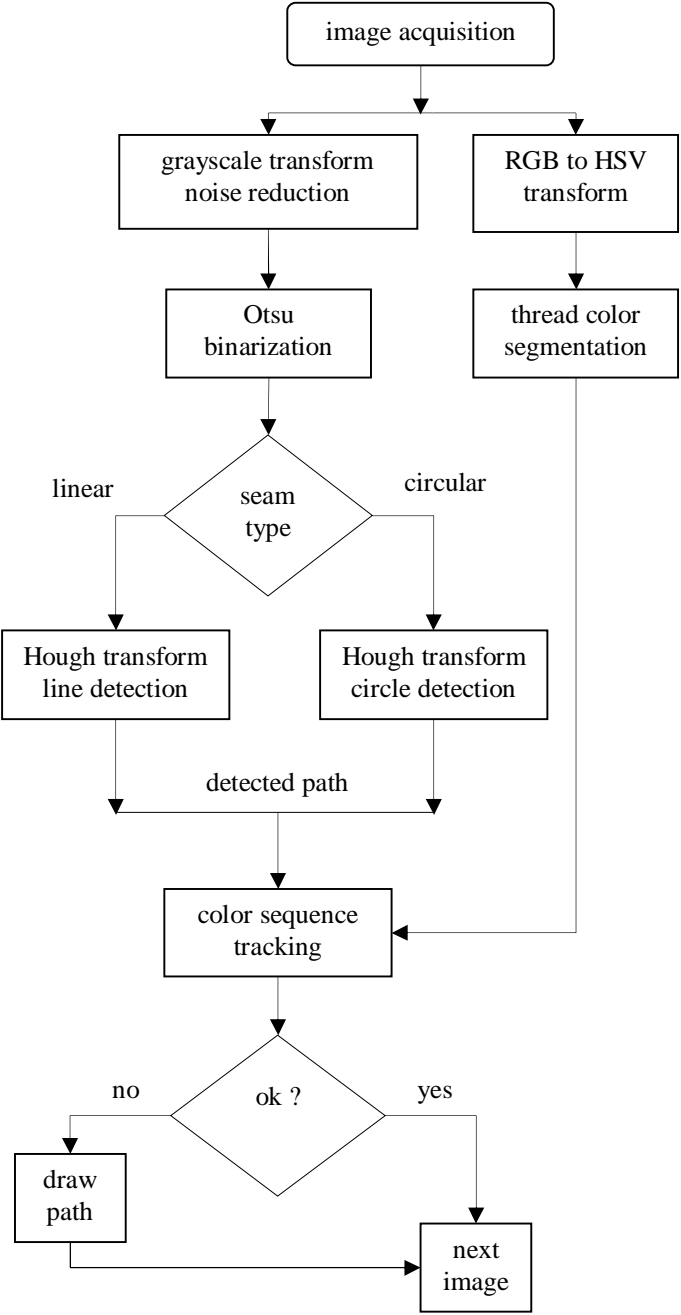

Figure 1. Flowchart of image processing.

In view of linear or circular path recognition, the image is converted from color to a grayscale format and smoothed for noise attenuation, using a Gaussian filter. The image binarization was performed using a version of the Otsu algorithm, which assumes that each of the two classes (object and background) is characterized by a range of intensities. Thus, the image histogram is bimodal and the optimal threshold will separate the two classes of pixels [22]. This process involves iteration for all possible threshold values, creating classes of objects (foreground) and background. The optimal threshold minimizes the inter-class variance, defined as the weighted sum of the variances of each class (equation 1).

$$s_w^2(t) = w_1(t)s_1^2(t) + w_2(t)s_2^2(t) \tag{1}$$

where $t$ is the threshold and $w_i(t), i = \overline{1,2}$ are the class probabilities estimated as:

$$w_1(t) = \sum_{i=1}^{t} p(i) \text{ and } w_2(t) = \sum_{i=t+1}^{I} p(i) \tag{2}$$

and the average of each class are given by:

$$m_1(t) = \sum_{i=1}^{t} \frac{ip(i)}{w_1(t)} \text{ and } m_2(t) = \sum_{i=t+1}^{I} \frac{ip(i)}{w_2(t)} \tag{3}$$

where the individual class variances are:

$$s_1^2(t) = \sum_{i=1}^{t} [i - m_1(t)]^2 \frac{p(i)}{w_1(t)} \text{ and } s_2^2(t) = \sum_{i=t+1}^{I} [i - m_2(t)]^2 \frac{p(i)}{w_2(t)} \tag{3}$$

The detection of linear seams is made using the Hough transform [23]. The basic idea behind, is to detect straight lines in an image assessing the pixels position and their colinearity. All the pixels belonging to the object class, from the previous stage of binarization will be considered based on their coordinates. It allows passing from the Cartesian coordinate system to the polar system characterized by the distance $\rho$ between the line and the origin and the angle $q$ of the vector from the origin to the nearest point of the line, using the equation:

$$r = x\cos q + y\sin q \tag{4}$$

Each line in the Cartesian space will have a unique pair of coordinates ($\rho$, $q$) in the Hough space. Therefore, a point of coordinates ($x_0$, $y_0$), all the lines passing through it will determine an infinity of pairs ($\rho$, $q$), arranged on a sinusoidal curve in the Hough space. A set of points that forms a line in the Cartesian space will produce a set of sinusoids intersecting in the point establishing the line parameters.

Using pairs of binarized pixels, the transform is applied for the area of interest and the intersections of sinusoids in Hough space are discovered. To retain the information about intersections, a data structure called accumulator, will be used. It records the intersections coordinates and the number of sinusoids superimposed. Finally, the accumulator will be inspected and intersection numbers bigger than a predetermined threshold are searched and converted into the Cartesian space to be displayed as lines.

The detection of circles is useful in determining the circular contour seams, therefore, the Hough transform for circles uses the same accumulator structure to retain information regarding the detected circle center [24]. Iteratively, for each pixel belonging to the structure binarized in the previous step, a circle of radius *R* is drawn in the accumulator (a two-dimensional array of image size). The local maxima represent the cells with highest drawn circle intersections, consequently the center of the detected circle of radius *R*. A priori knowledge of the circular seams radius will require less iteration for *R*. As a result, detecting the center and the radius of the circle will describe the circular seam.

As presented in the previous section, colored threads are used for the airbags seams. In view of a sewing tension control and monitoring and stitch line detection, color segmentation has to be used. As the image is acquired in RGB mode, precise color segmentation was performed in HSV color space. The corresponding red and green values for the threads are characterized by precise Hue and Saturation parameters and were employed in the classification of pixels. Colors are scanned on a local neighborhood sized 5x1, perpendicular and along the previously detected shape, following three types of patterns, corresponding to the nonconformities to be processed. If the rule is not respected, the pixels corresponding to the linear path are displayed in green, while red boxes mark the missing pattern for the circular seam.

## 4. Results

The algorithm for linear or circular seams control is analyzing the colors being present in the acquired image along the recognized lines or arcs by the Hough transform. A sequence of color appearances will be pursued and if an incorrect one is detected, a defective seam region will be revealed.

The image processing stages of straight seams detection are presented in figure 2. First, two operations will take place: an image conversion to gray levels and noise removal by smoothing, in order to diminish the sensor noise. The next step will be the image binarization using the Otsu threshold. In this case, the object of interest is the seam and will be displayed with black pixels, while the fabric will be labeled as background (figure 2 b, with a computed threshold value of 81). After thresholding binarization, the Hough transform will detect lines in the image as the seams should be straight and complete (figure 2 d, using a specified threshold of 55 for the accumulator).

The color image segmentation has the purpose of detecting correct formed stitches. In this respect, a transformation in the HSV color space will be performed, followed by a segmentation using descriptors for the tracked colors. The evaluation will consist of a color sequence inspection. If the sequence is not correct, a negative diagnosis result will be displayed as the seam is either incomplete or broken, either the direction is deviated.

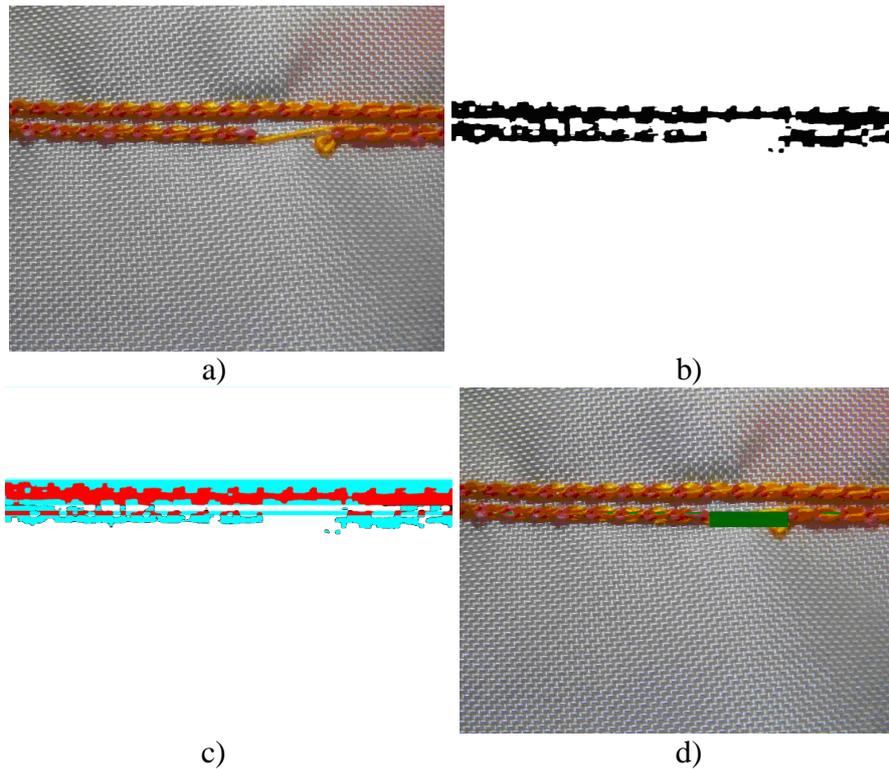

Figure 2. Original image of an airbag double seam with a) missing stitches; b) binarized image; c) color segmentation; d) defect detection result.

The assessing of circular or curved seams is based on the arcs/circle recognition using the corresponding Hough transform (figure 3). The detection of the circular outline of the stitching is performed in two steps: first, the estimated calculations of the radius of the circle determined by seam, and then the calculating of the center using the circle equation. The stitches are represented by binarized pixels located on predefined shapes. Further more, a neighboring block scans along the detected form and searches for a color sequence in the HSV space (figure 3 b). The presence of two colors per stitch is checked and points out an incorrect seam (figure 3 c).

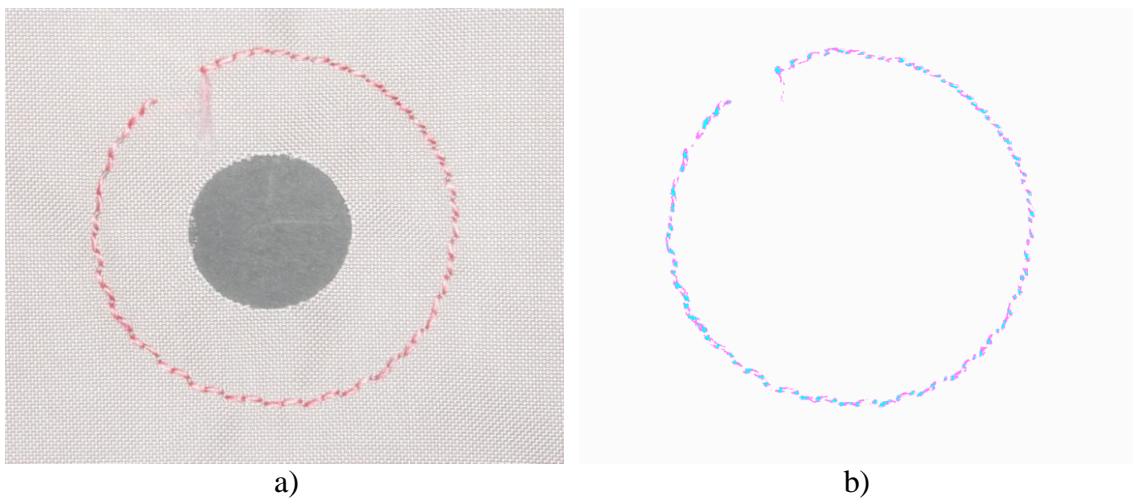

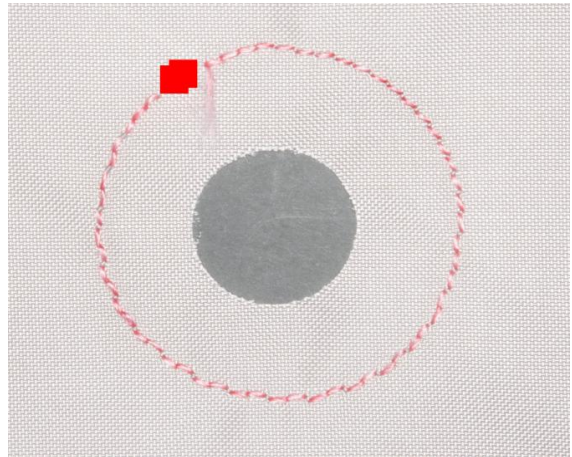

c)

Figure 3. Circular seam with thread missing stitches, a) vent reinforcement seam image, b) color segmentation, c) defect detection.

Using this type of fabric, two tests were made, both being diagnosed as faulty seams. For the case presented in figure 3, the defect consists of a seam thread break, which means incomplete stitching. The lack of seam is marked correspondingly. In figure 4, the circular seam is superimposed, therefore, the defect is symbolized with red marks. Another example of linear skipped stitch detection using the color segmentation is presented in figure 5.

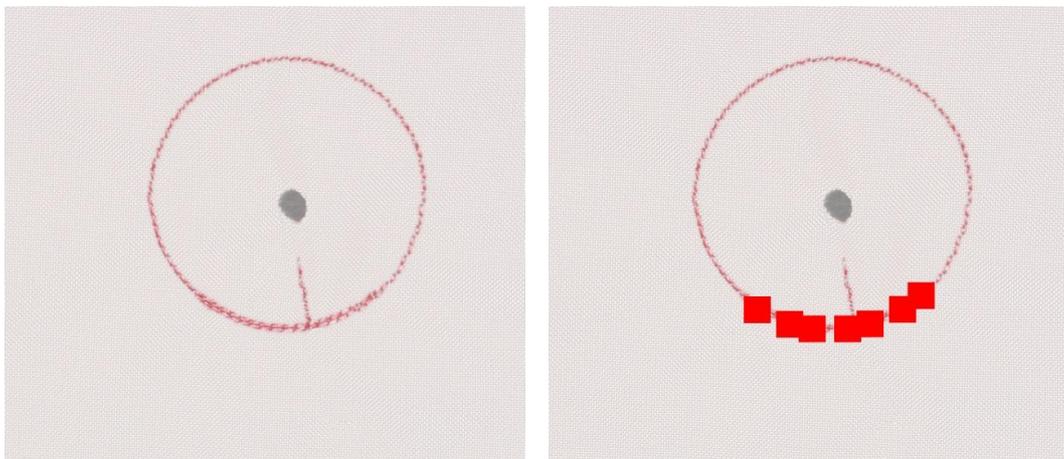

a)                                           b)

Figure 4. A long superimposed circular seam, a) original image, b) detected nonconformity.

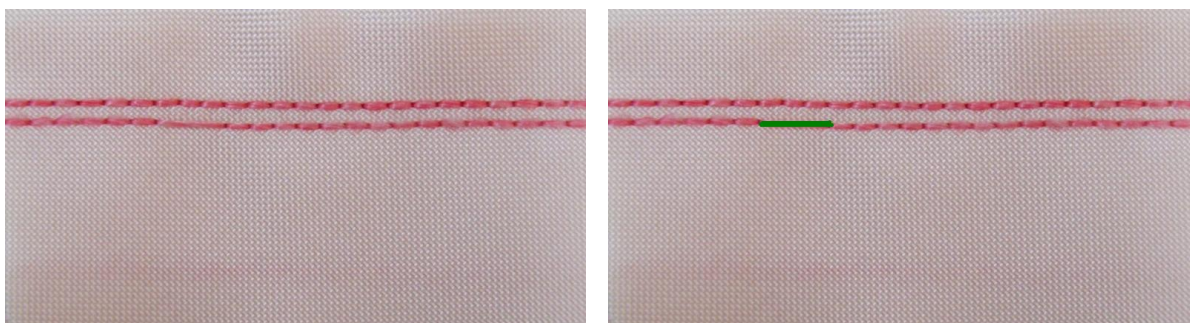

a)                                           b)

Figure 5. Skipped stitch detection, a) input image, b) detected seam.

## 5. Conclusions

The airbag, jointly with the seat belts, must assure passengers safety in the case of a car accident. Besides the properties of fabric, the assembly seams play an important role in airbag functionality. Due to the fact that the seams cannot be repaired during the production process, any stitching defect will cause a nonconforming product. Intermediate inspection of stitches before the final closure of the airbag is an important stage, since it prevents non-compliant subassembly to pass on the next production step.

In the airbag sewing stage, fabric and reinforcements parts are transformed into a final product using threads of different linear densities and colors. Seams paths can be linear or circular (including arcs and combined forms).

The process of automatic defect detection depend on the seams paths, therefore two different methods were employed in the cases of straight and circular seams and the most common defects of sewing were considered: missed stitch, skipped and superimposed stitches. Using a grayscale combined with a color image processing, the correct rule of seam formation is checked along the detected paths and the corresponding nonconformities are highlighted.

The processing framework succeeded to inspect correctly all the collected defects. Automatic defects detection of seams keeps the sewing process in control and makes the final product comply with the client requirements.

**Acknowledgment**

The Authors would like to acknowledge TAKATA PETRI Sibiu, for providing the samples used in this study.

**Conflict of Interests**

The authors declare that there is no conflict of interests regarding the publication of this paper.